# Simple Recurrent Neural Networks is all we need for clinical events predictions using EHR data

Laila Rasmy [a], Jie Zhu [b], Zhiheng Li [c], Xin Hao [d], Hong Thoai Tran [b], Yujia Zhou [a], Firat Tiryaki [a], Yang Xiang [a], Hua Xu [a], Degui Zhi [a]

[a] *School of Biomedical Informatics, University of Texas Health Science Center at Houston, Texas, USA,*
[b] *School of Public Health, University of Texas Health Science Center at Houston, Texas, USA,*
[c] *School of Computer Science and Technology, Dalian University of Technology, Dalian, China,*
[d] *Computer Science Department, School of Engineering, Rice University, Texas, USA,*

## Introduction

Recently it is of great interest to investigate the application of deep learning models for the prediction of clinical events using electronic health records (EHR) data. In EHR data, the longitudinal history of a patient is often represented as sequence data, where each patient can be represented as a sequence of visits, and each visit contains multiple events. As a result, deep learning models developed for sequence modeling, especially recurrent neural networks (RNNs), were developed for EHR-based clinical events prediction [1].

While a large variety of candidate RNN models were proposed, it is unclear if complex architecture innovations will offer higher predictive performance. In order to move this field forward, a rigorous evaluation of various methods is needed. For example, a thorough comparison of RNN models in natural language processing (NLP) models found that simple RNN models often offer competitive results when properly tuned [2,3] In this study, we aim to offer a similarly comprehensive evaluation of RNN-based methods in modeling EHR data for predicting patient risk for developing heart failure or early readmit for an inpatient hospitalization.

## Methods

RNN models vary either through the cell architecture or the connection between the cells. we tested 3 cells variations namely vanilla RNN cell with tanh activation, long short term memory (LSTM), and gated recurrent unit (GRU), and 3 different connections, Standard, Bidirectional and Dilated connections[4]. We also tested two recent models, Quasi-RNN (QRNN)[5] and Time aware LSTM (T-LSTM) [6], besides RETAIN [7]. In addition to these 12 variations of RNN based models, we also included non-deep learning methods, Random forest (RF) and logistic regression (LR) for comparison. For each RNN model architecture, we experimented on 7 optimizers (Adam, Adamax, Adagrad, Adadelta, RMSprop, ASGD, and SGD), and used Bayesian optimization [8] for searching hyperparameters, including embedding dimension, hidden size, learning rate, weight decay, and numerical tolerance. All our implementations used PyTorch v0.4 framework. All our cohorts were split into train, validation and test sets on the 7:1:2 ratios. We used the area under the ROC curve (AUROC) as our main evaluation criteria.

We tested the performance of different RNN models on two tasks. The first is the prediction of patient risk to develop Heart Failure (HF) by the next visit as an example of a clinical disease-specific question. The second is to predict the risk of early readmission for patients with HF, Myocardial infarction (MI), Stroke (CVD), Pneumonia (Pn) and Chronic Obstructive Pulmonary Disease (COPD) diseases publicly reported by CMS, as an example of healthcare quality of service related question. We extracted cohorts for both tasks from the Cerner Healthfacts® database (2016 version).

For the Heart Failure cohort, we used hospital_5 data described in [9] with a cohort of 5,010 cases and 37,719 controls. For the Readmission cohort (Readm), we used the same hospital data and extracted patients with the five diseases mentioned above. We used the diagnosis, medication, and encounter administrative information for a patient as input. We used the time differences between two consecutive in-patient encounters excluding all recurring or transfer encounters to define our cohort. We defined cases as patients readmitted before 30 days and for better discrimination, our control group consists of patients readmitted after 90 days. For one hospital, we obtained 5,897 cases and 4,757 controls.

## Results and Discussions

We trained a large number of models with hyperparameter variations on each cohort. We identified the best model within a specific category based on the validation AUC. Figure 1 is showing the AUC on the Test set. Interestingly, the best performing model was the simple unidirectional GRU. In addition, we tested ensembling GRU with LR but that does not improve the model. Overall, we found our results inline with NLP researchers observations[2,3]. Baseline RNN models such as GRU are often sufficient for predictive modeling tasks in EHR. One of the limitations of the current study is that we focused on a single hospital, due to constraints of computational resources. We plan to test and compare on the full cohort that includes hundreds of thousands of patients across multiple hospitals to further verify our conclusions.

*Table 1 - Benchmarking Results*

| Model | HF | Readm | Model | HF | Readm |
|---|---|---|---|---|---|
| **GRU** | 84.8 | 75.5 | **Bi-GRU** | 84.5 | 74.4 |
| **LSTM** | 83.9 | 73.8 | **Bi-LSTM** | 84.4 | 75.2 |
| **D-GRU** | 83.3 | 73.5 | **Bi-RNN** | 83.1 | 74.1 |
| **D-LSTM** | 83.3 | 72.8 | **D-RNN** | 83.2 | 70.9 |
| **QRNN** | 83.2 | 71.5 | **Vanilla-RNN** | 83.3 | 63.9 |
| **RETAIN** | 83.8 | 70.1 | **RF** | 78.8 | 73.6 |
| **LR** | 79.0 | 67.0 | | | |